\documentclass[10pt,twocolumn,letterpaper]{article}

\usepackage{iccv}
\usepackage{times}
\usepackage{epsfig}
\usepackage{graphicx}
\usepackage{amsmath}
\usepackage{amssymb}

\usepackage{dblfloatfix} 
\usepackage{subcaption}
\usepackage{pifont}
\usepackage{mathtools}

\DeclarePairedDelimiter\abs{\lvert}{\rvert}\DeclarePairedDelimiter\norm{\lVert}{\rVert}

\makeatletter
\let\oldabs\abs
\def\abs{\@ifstar{\oldabs}{\oldabs*}}
\let\oldnorm\norm
\def\norm{\@ifstar{\oldnorm}{\oldnorm*}}
\makeatother

\newcommand{\xmark}{\ding{55}}
  \DeclareMathOperator{\sgn}{sgn}

\usepackage[pagebackref=true,breaklinks=true,colorlinks,bookmarks=false]{hyperref}

\iccvfinalcopy 

\ificcvfinal\fi
\begin{document}

\title{Improved training of binary networks for human pose estimation and image recognition}

\author{Adrian Bulat  \qquad Georgios Tzimiropoulos \qquad Jean Kossaifi \qquad Maja Pantic
        \vspace{5pt}\\
		Samsung AI Center, Cambridge\\
		United Kingdom\\
		{\tt\small \{adrian.bulat, georgios.t, j.kossaifi, maja.pantic\}@samsung.com}
		}

\maketitle

\begin{abstract}

Big neural networks trained on large datasets have advanced the state-of-the-art for a large variety of challenging problems, improving performance by a large margin. However, under low memory and limited computational power constraints, the accuracy on the same problems drops considerable. In this paper, we propose a series of techniques that significantly improve the accuracy of binarized neural networks (i.e networks where both the features and the weights are binary). We evaluate the proposed improvements on two diverse tasks: fine-grained recognition (human pose estimation) and large-scale image recognition (ImageNet classification). 
Specifically, we introduce a series of novel methodological changes including: (a) more appropriate activation functions, (b) reverse-order initialization, (c) progressive quantization, and (d) network stacking and show that these additions improve existing state-of-the-art network binarization techniques, significantly. Additionally, for the first time, we also investigate the extent to which network binarization and knowledge distillation can be combined. When tested on the challenging MPII dataset, our method shows a performance improvement of more than 4\% in absolute terms. Finally, we further validate our findings by applying the proposed techniques for large-scale object recognition on the Imagenet dataset, on which we report a reduction of error rate by 4\%.

\end{abstract}

\section{Introduction}

Recent methods based on Convolutional Neural Networks (CNNs)
have been shown to produce results of high accuracy for a wide range of challenging Computer Vision tasks like image recognition \cite{krizhevsky2012imagenet, simonyan2014very,he2016deep}, object detection \cite{ren2015faster}, semantic segmentation \cite{long2015fully, he2017mask} and human pose estimation \cite{xiao2018simple, newell2016stacked}. Two fundamental assumptions made by these methods are that: 1) very large and diverse labelled datasets are available for training, and 2) that at least one high-end GPU is available for model training and inference. While it can be assumed that for training both labelled training data and computational resources are available, from a a practical perspective, in many applications (e.g. object recognition or human sensing on mobile devices and robots), it is unreasonable to assume that dedicated high-end GPUs are available for inference. The aim of this paper is to enable highly accurate and efficient convolutional networks on devices with limited memory, storage and computational power. Under such constraints, the accuracy and performance of existing methods rapidly drops, and the problem is considered far from being solved.

Perhaps, the most promising method for model compression and efficient model inference is network binarization, especially when both activations and weights are binary  \cite{courbariaux2015binaryconnect, courbariaux2016binarized, rastegari2016xnor}. In this case, the binary convolution operation can be efficiently implemented with the bitwise \textit{XNOR}, resulting in speed-up of $\sim 58\times$ on CPU (this speed-up on FPGAs can be even higher) and model compression ratio of $\sim 32\times$~\cite{rastegari2016xnor}. Although no other technique can achieve such impressive speed-ups and compression rates, this also comes at the cost of reduced accuracy. For example, there is $ \sim 18 \%$ difference in top-1 accuracy between a real-valued ResNet-18 and its binary counterpart on ImageNet~\cite{rastegari2016xnor}, and $ \sim 9 \%$ difference between a real-valued state-of-the-art network for human pose estimation and its binary counterpart on MPII \cite{bulat2017binarized}.

Motivated by the above findings, in this work, we focus on improving the training of binary networks by proposing a series of methodological improvements. In particular, we make the following \textbf{contributions}: 
\begin{itemize}
    \item 
    We motivate, provide convincing evidence and describe a series of methodological changes for training binary neural networks including (a) more appropriate non-linear activation functions (Sub-section~\ref{ssec:Symmetrical}), (b) reverse-order initialization  (Sub-section~\ref{ssec:progressive-initialisation}), (c) progressive quantization (Sub-section~\ref{ssec:progressive-quantization}), and (d) network stacking (Sub-section~ \ref{ssec:stacked-binary-nets}) that, individually and combined, are shown to improve existing state-of-the-art network binarization techniques, significantly. (e) We also show to what extent network binarization and knowledge distillation can be combined (Section~\ref{sec:distilled-binary-nets}).
    \item 
    We show that our improved training of binary networks is task and network agnostic by applying it on two diverse tasks: fine-grained recognition and, in particular, human pose estimation and classification, specifically ImageNet classification.
     \item 
    Exhaustive experiments conducted on the challenging MPII dataset show that our method offers an improvement of more than 4\% in absolute terms over the state-of-the-art (Section~\ref{sec:results-human-pose}). 
    \item 
    On ImageNet we report a reduction of error rate by 4\%  over the current state-of-the-art (Section~\ref{sec:results-imagenet}). 
\end{itemize}

\section{Related work}\label{sec:related-work}
In this section, we review related prior work including network quantization and knowledge distillation for image classification, and methods for efficient human pose estimation.

\subsection{Network Quantization}\label{ssec:related-network-quantization}

Network quantization refers to quantizing the weights and/or the features of a neural network. It is considered the method of choice for model compression and efficient model inference and a very active topic of research. Seminal work in this area goes back to \cite{courbariaux2014training, lin2015fixed} who introduced techniques for 16- and 8-bit quantization. The method of \cite{zhou2016dorefa} proposed a technique which allocates different numbers of bits (1-2-6) for the network parameters, activations and gradients. For more recent work see \cite{tung2018clip,wu2016quantized,zhou2018explicit,tang2018quantized}.

The focus of the first methods proposed in this work is on binarization of both weights and features which is the extreme case aiming to quantizing to $\{-1,1\}$, thus offering the largest possible compression and speed gains. The work of \cite{courbariaux2015binaryconnect} introduced a technique for training a CNN with binary weights. A follow-up work \cite{courbariaux2016binarized} demonstrates how to binarize both parameters and activations. This has the advantage that, during the forward pass, multiplications can be replaced with binary operations. The method of \cite{rastegari2016xnor} proposes to model the weights with binary numbers multiplied by a scaling factor. Using this simple modification which does not sacrifice the beneficial properties of binary networks, \cite{rastegari2016xnor} was the first to report good results on a large scale dataset (ImageNet~\cite{deng2009imagenet}). 

Our method proposes several extensions to \cite{rastegari2016xnor}, including more appropriate activation functions, reverse-order initialization, progressive quantization, and network stacking, which are shown to produce large improvements of more than \% 4 (in absolute terms) for human pose estimation over the state-of-the art \cite{bulat2017binarized}. We also report similar improvements for large-scale image classification on ImageNet, in particular, we report a reduction of error rate by 4\% over the current state-of-the-art \cite{rastegari2016xnor}.

\subsection{Knowledge Distillation}\label{ssec:related-knowledge distillation}
Recent works~\cite{hinton2015distilling} have shown that, at least for real-valued networks, the performance of a smaller network can be improved by ``distilling the knowledge'' of another one, where ``knowledge distillation'' refers to transferring knowledge from one CNN (the so-called ``teacher'') to another (the so-called ``student''). Typically, the teacher is a high-capacity model of great accuracy, while the student is a compact model with much fewer parameters (thus also requiring much less computation).
Thus, the goal of knowledge distillation is to use the teacher to train a compact student model with similar accuracy to that of the teacher. The term ``knowledge'' refers to the soft outputs of the teacher. Such soft outputs provide extra supervisory signals of intra-class and inter-class similarities learned by teacher. Further extensions include transferring from intermediate representations of the teacher network \cite{romero2015fitnets} and from attention maps \cite{zagoruyko2017paying}. While most of the prior work focuses on distilling real-valued neural networks little to no-work has been done on studying the effectiveness of such approaches for binarized neural networks. In this work, we propose to adapt such techniques to binary networks, showing through empirical evidence their positive effect on improving accuracy.

\subsection{Human pose estimation}\label{ssec:related-human-pose}

A large number of works have been recently proposed for both single-person \cite{newell2016stacked,wei2016convolutional,bulat2016human,ke2018multi,tang2018deeply,chu2017multi,yang2017learning,chen2017adversarial} and multi-person \cite{cao2017realtime, papandreou2017towards, he2017mask, girdhar2018detect} human pose estimation. We note that the primary focus of these works is accuracy (especially for the single-person case) rather than efficient inference under low memory and computational constraints which is the main focus of our work.

Many of the aforementioned methods use the so-called HourGlass (HG) architecture \cite{newell2016stacked} and its variants. 
While we also used the HG in our work, our focus was to enhance its efficiency while maintaining as much as possible its high accuracy which makes our work different to all aforementioned works. To our knowledge, the only papers that have similar aims are the works of \cite{bulat2017binarized}  and \cite{tang2018quantized}.
\cite{bulat2017binarized} and its extension \cite{bulat2018hierarchical} aim to improve binary neural networks for human pose estimation by introducing a novel residual block. \cite{tang2018quantized} aims to improve quantized neural networks by introducing a new HG architecture. In contrast, in this work, we focus on improving binary networks for human pose estimation by (a) improving the binarization process per se, and (b) combining binarization with knowledge distillation. Our method is more general than the improvements proposed in \cite{bulat2017binarized} and \cite{tang2018quantized}. We illustrate this by showing the benefits of the proposed method for also improving ImageNet classification with binary networks.

\begin{figure}[!htbp]

    \centering
    \includegraphics[width=1\linewidth]{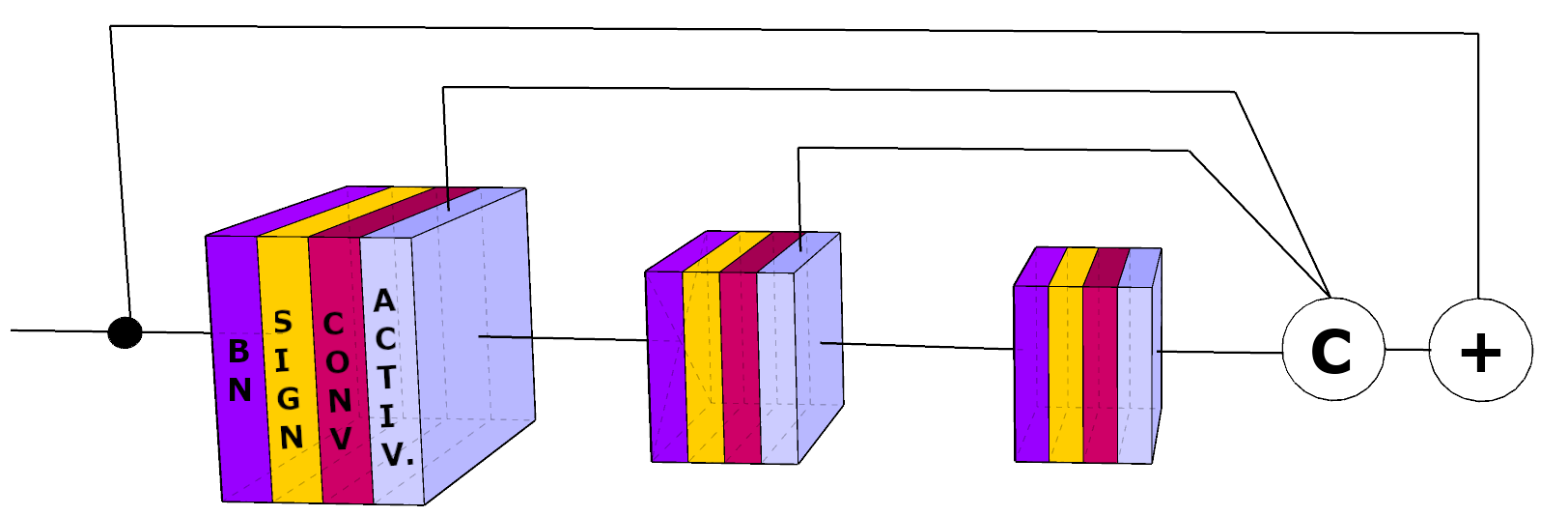}
    \caption{The residual binary block of~\cite{bulat2018hierarchical} used in our work. The module has a hierarchical, parallel, multi-scale structure comprised of three $3\times3$ convolutional layers with input-output channel ratio equal to 1:2, 1:4 and 1:4. Each convolution layer is preceded by a BatchNorm and the binarization function ($\sgn(x)$) and followed by a non-linearity. See Section \ref{sec:binarization} for the changes introduced in our work for improving its performance.} \label{fig:multi-scale-block}

\end{figure}

\section{Method}\label{sec:binarization}

This section presents the proposed methodological changes for improving the network binarization process. Throughout this section, we validated the performance gains offered by our method on the single person human pose estimation dataset, MPII. We note that we chose human pose estimation as the main dataset to report the bulk of our results as the dataset is considerably smaller and training is much faster (compared to ImageNet).  
 
Sub-section~\ref{ssec:baseline} describes the strong baseline used in our work, briefly explaining the binarization process proposed in\cite{rastegari2016xnor} and~\cite{bulat2017binarized}, while the proposed improvements are described in Sub-sections~\ref{ssec:Symmetrical},  \ref{ssec:progressive-initialisation}, \ref{ssec:progressive-quantization}, \ref{ssec:stacked-binary-nets} and \ref{sec:distilled-binary-nets}.

\subsection{Baseline}\label{ssec:baseline}
All results reported herein are against the state-of-the-art method of~\cite{bulat2017binarized} which we used as a strong baseline to report the performance improvements introduced by our method. The method of~\cite{bulat2017binarized} combines the HourGlass (HG) architecture of~\cite{newell2016stacked} with the a newly proposed residual block that was specifically designed for binary CNNs (see Fig.~\ref{fig:multi-scale-block}). The network was binarized using the approach described in \cite{rastegari2016xnor} as follows:
\begin{equation}
    \mathbf{T} * \mathbf{W} \approx (\sgn(\mathbf{T}) \circledast \sgn(\mathbf{W}))\mathbf{K}\alpha   \label{eq:additive0},
\end{equation}
where $\mathbf{T}$ is the input tensor, $\mathbf{W}$ is the layer's weight tensor, $\mathbf{K}$ a matrix containing the scaling factors for all the sub-tensors of $\mathbf{T}$, and $\alpha \in \mathbb{R}^{+}$ is a scaling factor for the weights. $\circledast$ denotes the binary convolution operation which can be efficiently implemented with the bitwise \textit{XNOR}, resulting in speed-up of $\approx 58\times$ and model compression ratio of $\approx 32\times$~\cite{rastegari2016xnor}. Note than in practice, we follow~\cite{bulat2017binarized,rastegari2016xnor} and drop $\mathbf{K}$ since this speed-ups the network at a negligible performance drop. 

\subsection{Leaky non-linearities} \label{ssec:Symmetrical}

\begin{figure*}[!htbp]
    \centering
    \includegraphics[width=0.85\linewidth]{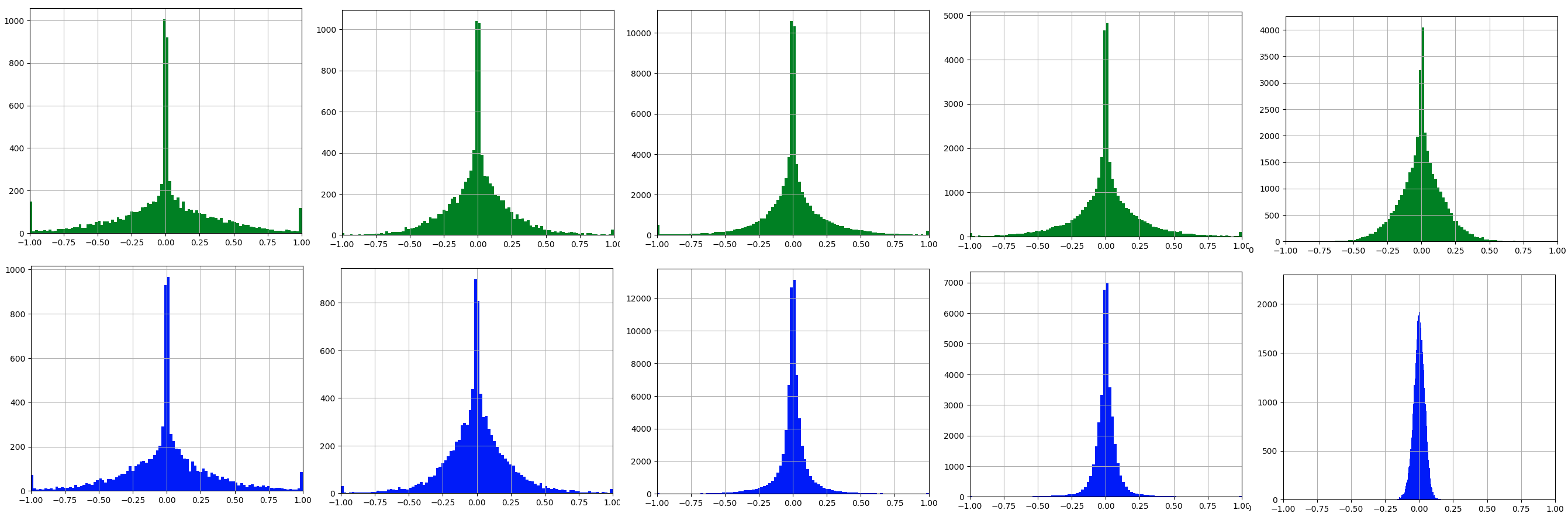}
    \caption{Weight distribution for various layers from a network using PReLU (first row) and ReLU (second row) as we advance in the network (from left to right). The ReLU tends to push the weights closer to 0 making a jump between states more likely, thus causing the observed instabilities.}
    \label{fig:weights-distribution}
\end{figure*}

Previous work~\cite{rastegari2016xnor,bulat2017binarized} has shown that adding a non-linearity after each convolutional layer can be used to increase the  performance of binarized CNNs. In the context of real-valued networks there exists a plethora or works that explore their effect on the overall network accuracy, however, in contrast to this, there is little to no work avaialble for binary networks.  Herein, we rigorously explore the choice of the non-linearity and its impact on the overall performance for the task of human pose estimation showing empirically in the process the negative impact of the previously proposed ReLU. Instead of using a ReLU, we propose to use the recently introduced PReLU~\cite{he2015delving} function, an adaptation of the leaky ReLU that has a learnable negative slope,  which we find it to perform better than both the ReLU and the leaky ReLU. 

\begin{figure}[!htbp]
    \centering
    \includegraphics[width=0.7\linewidth]{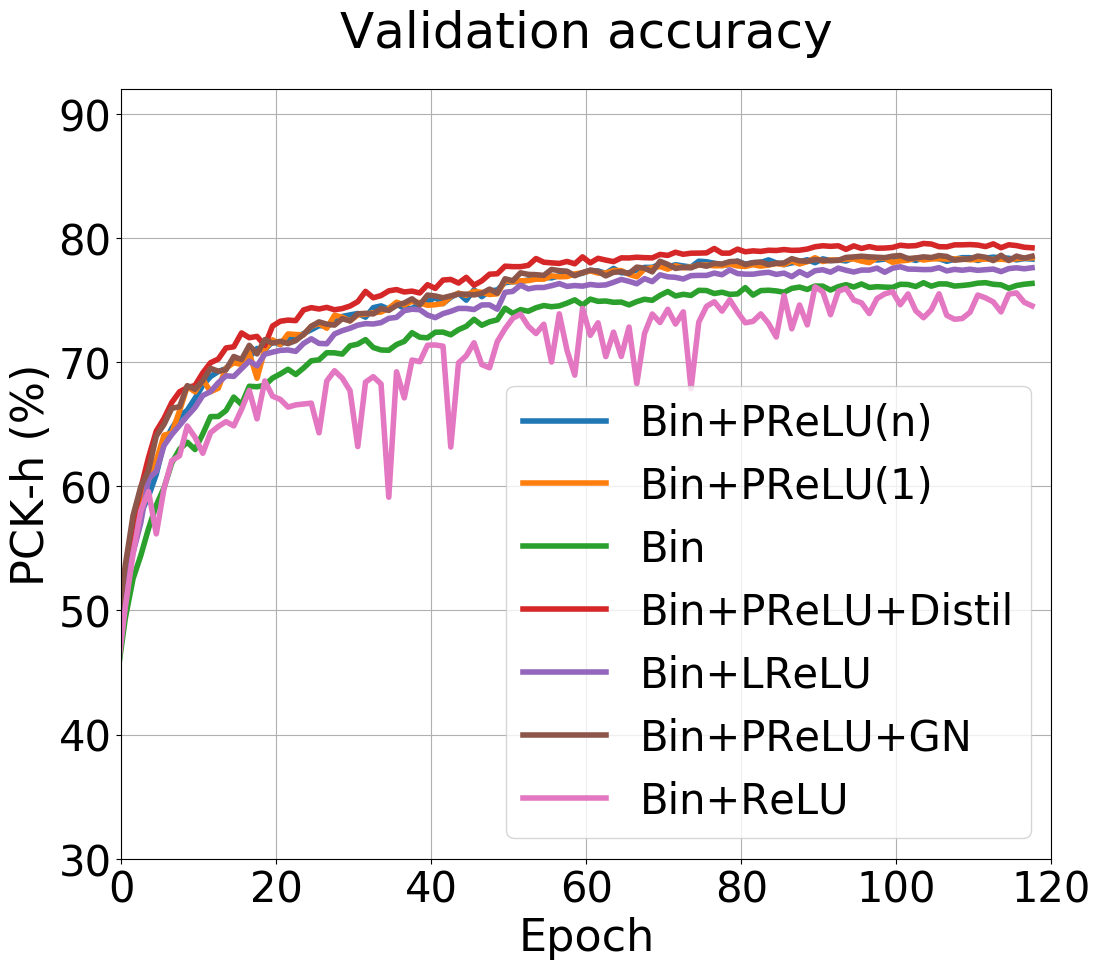}
    \caption{Accuracy evolution on the validation set of MPII during training. Notice the high fluctuations introduced by the ReLU. Best performance is obtained with PReLU.}
    \label{fig:validation-accuracy-main}
\end{figure}

There are two main arguments for justifying our findings. Firstly, with the help of the $\sgn$ function, the binarization process restricts the possible states of the filters and features to $\{-1, 1\}$. As such, the representational power of the network resides on these two states, and removing one of them during training using a ReLU for each convolutional layer makes the training unstable. See also Fig.~\ref{fig:validation-accuracy-main}. Secondly, this instability is further amplified  by the fact that the implementation of the sign function is ``leaky" at 0, introducing a third unwanted spurious state and the subsequent iterations can cause easy jumps between the two states. See also Fig. \ref{fig:weights-distribution}. Note, that despite the fact that the Batch Normalisation~\cite{ioffe2015batch} layer mitigates some of this effects by re-centering the input distribution, as the experiments show, in practice, the network can achieve significantly better accuracy if the non-linearity function allows negative values to pass. On the other hand, we know that non-linearities should be used to increase the representational power of the network. We conclude that a PReLU can be safely used for this purpose removing also the aforementioned instabilities.

\subsection{Reverse-order initialization}\label{ssec:progressive-initialisation}

Initialization of neural networks has been the subject of study of many recent works~\cite{glorot2010understanding,he2015delving,saxe2013exact} where it was shown that an appropriate initialization is often required for achieving good performance~\cite{sutskever2013importance}. The same holds for quantized networks, where most of prior works either use an adaptation of the above mentioned initialization strategies, or start from a pretrained real-valued neural network. However, while the weight binarization alone can be done with little to no accuracy loss~\cite{rastegari2016xnor}, quantizing the features has much higher detrimental effect~\cite{rastegari2016xnor,zhou2016dorefa}. In addition, since the output signal from $\sgn$ is very different to the output of a ReLU layer, the transition from a fully real-valued network to a binary one causes a catastrophic loss in accuracy often comparable with training from scratch.

\begin{figure}[!htbp]
    \centering
    \includegraphics[width=0.7\linewidth]{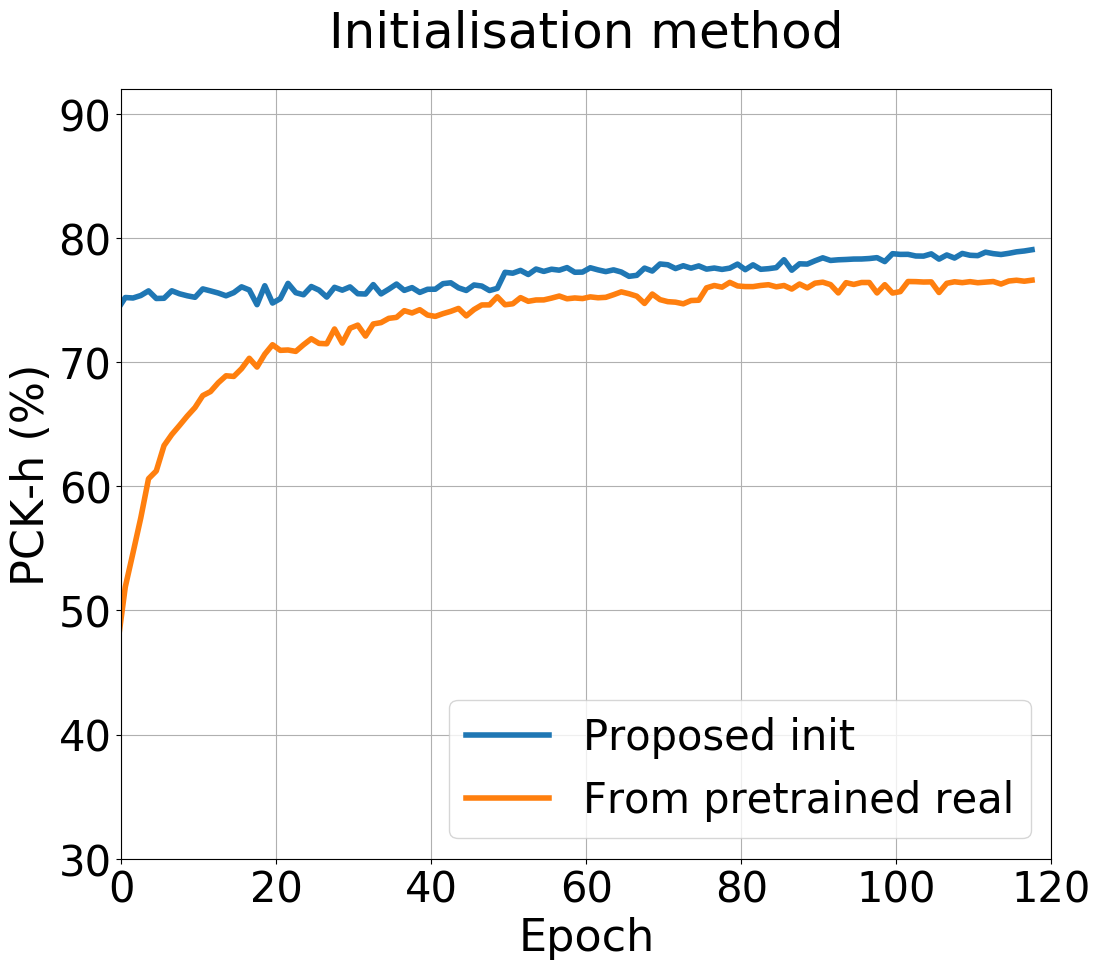}
    \caption{Accuracy evolution on the validation set of MPII during training for different pre-initialization approaches. Our initialization provides a much better starting point.}
    \label{fig:init-method}
\end{figure}

To alleviate this, we propose the opposite of what is currently considered the standard method to initialize a binary network from a real-valued one: we propose to firstly train a network with real weights and binary features (the features are binarized using the approach presented in Section~\ref{ssec:progressive-quantization}) and only after this, it is fully trained to further binarize the weights. By doing so, we effectively split the problem into two sub-problems: weight and feature binarization which we then try to solve from the hardest to the easiest one. Fig.~\ref{fig:init-method} shows the advantage of the proposed initialization method against standard pre-training. 

\subsection{Smooth progressive quantization}\label{ssec:progressive-quantization}

Previous works have shown that incrementally quantizing the network either by gradually decreasing the precision or by partitioning and progressively increasing the amount of quantized weights~\cite{zhou2017incremental} leads to decent performance improvements. While the later is more practical, it requires a careful fine-tuning of the quantization ratio at each step. 

Instead, in this work, we follow a different route by proposing to approximate the quantization function $\sgn(x)$ with a smoother one, in which the estimation error is controlled by $\lambda$. By gradually increasing $\lambda$ during training, we achieve a progressive binarization. This allows for a natural and smoother transition in which the selection of the weights to be binarized occurs implicitly and can be easily controlled by varying $\lambda$ without the need to define a fixed scheduling for increasing the amount of quantized weights as in~\cite{zhou2017incremental}. 

\begin{figure}[!htbp]
    \centering
    \includegraphics[width=1\linewidth]{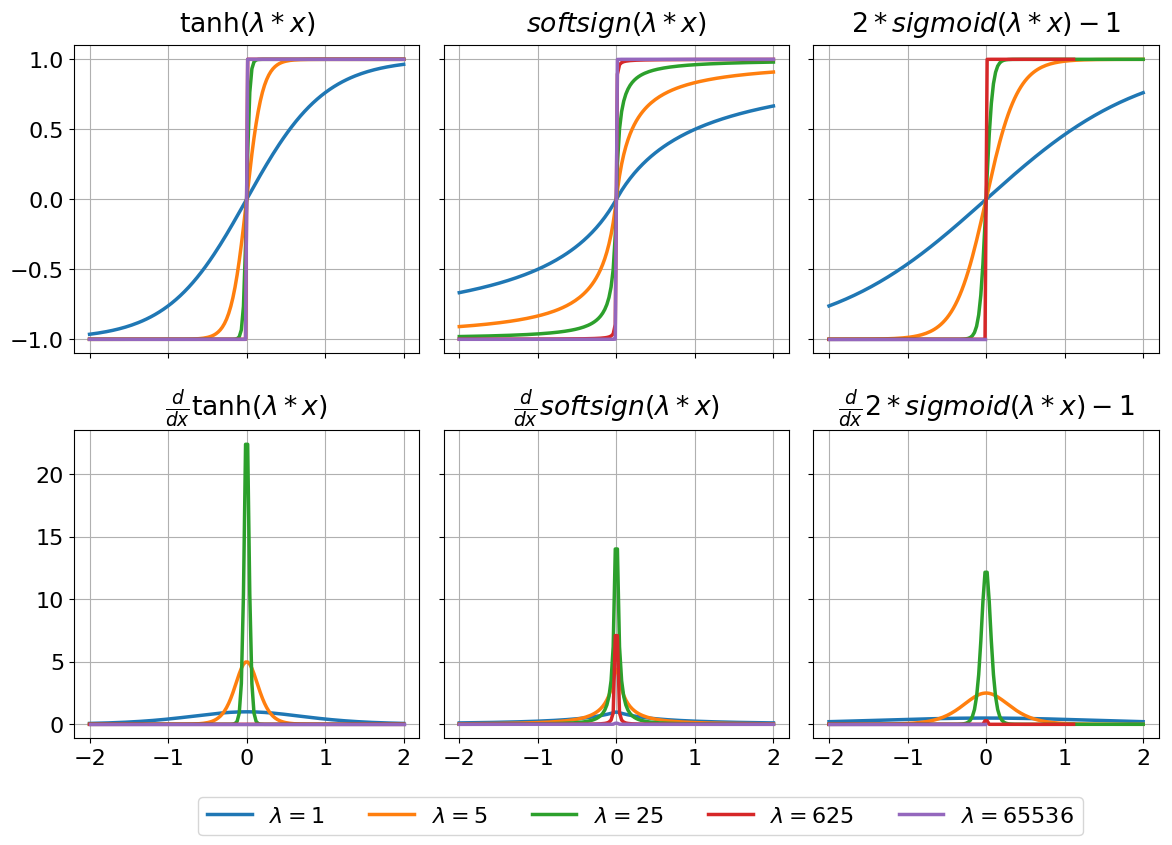}
    \caption{The quantization approximation functions used: sigmoid, softsign and tanh (first row) and their derivatives (second row) for various values of $\lambda=\{1,5,25,625,65536\}$.}
    \label{fig:functions-plot}
    \vspace{5pt}
\end{figure}

In the following, we present a few options we explored to approximate the $\sgn(x)$ function alongside their derivatives (see also Fig.~\ref{fig:functions-plot}):

\noindent
\textbf{Sigmoid:}
\begin{equation}
    \begin{aligned}
    \sgn(x) \approx 2\Big(\frac{e^{\lambda x}}{1+e^{\lambda x}}\Big)-1 \\
    \frac{d}{dx}2\Big(\frac{e^{\lambda x}}{1+e^{\lambda x}}\Big)-1 = \frac{2\lambda e^{\lambda x}}{(e^{\lambda x} +1 )^2}
    \end{aligned}
\end{equation}

\noindent
\textbf{SoftSign:}

\begin{equation}
    \begin{aligned}
    \sgn(x) \approx \frac{\lambda x}{1+\lambda \abs{x}} \\
    \frac{d}{dx}\frac{\lambda x}{1+\lambda \abs{x}} = \frac{\lambda}{(1+\lambda \abs{x})^2}
    \end{aligned}
\end{equation}

\noindent
\textbf{Tanh:}

\begin{equation}
    \begin{aligned}
    \sgn(x) \approx \tanh(\lambda x) \\
    \frac{d}{dx} \tanh(\lambda x) = \lambda(1-\tanh^2(\lambda x))
    \end{aligned}
\end{equation}

\begin{figure*}[!htbp]
    \centering
    \begin{subfigure}[t]{0.3\textwidth}
    \includegraphics[width=0.9\linewidth]{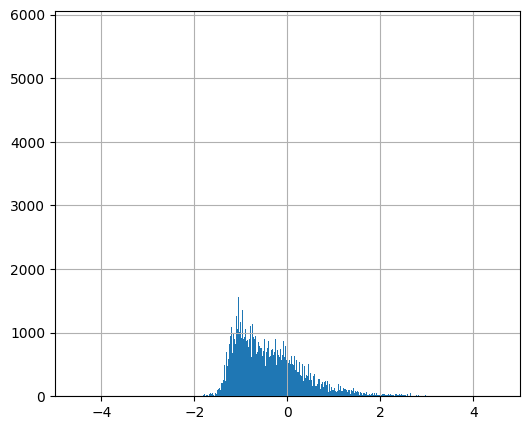}
    \label{fig:input-dist-1}
    \end{subfigure}
    ~
    \begin{subfigure}[t]{0.299\textwidth}
    \includegraphics[width=0.9\linewidth]{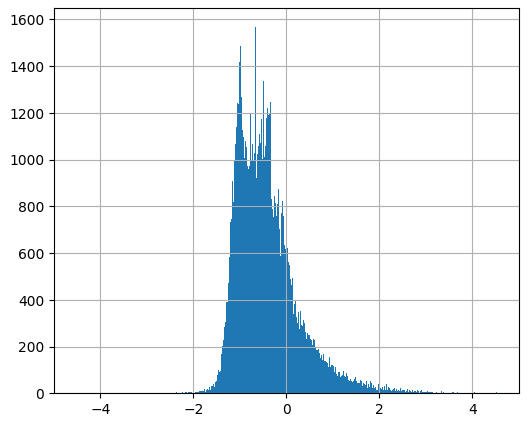}
    \label{fig:input-dist-2}
    \end{subfigure}
     ~
     \begin{subfigure}[t]{0.303\textwidth}
    \includegraphics[width=0.9\linewidth]{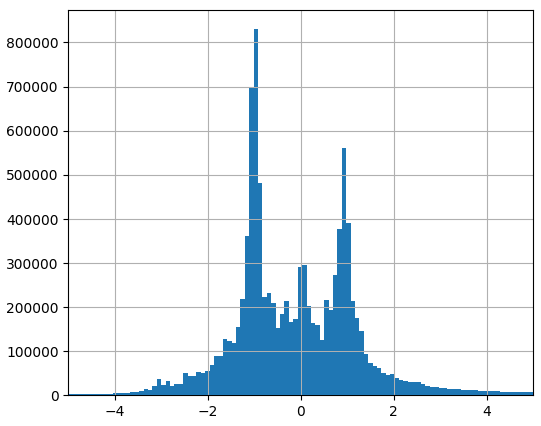}
    \label{fig:input-dist-3}
    \end{subfigure}
    
    \caption{Input distribution before the $\sgn$ function from 3 layers located at the bottom, middle and top of the network. Most values are in a range  where the approximation function outputs values close to $\pm{1}$ allowing the approximator to reach good estimates for relatively low values of $\lambda$.}
    \label{fig:input-distribution}
\end{figure*}

\begin{figure*}[!htbp]
    \begin{subfigure}[t]{0.24\textwidth}
    \centering
    \includegraphics[width=1\linewidth,  trim={0.5cm 0.5cm 0.5cm 0.5cm}, clip]{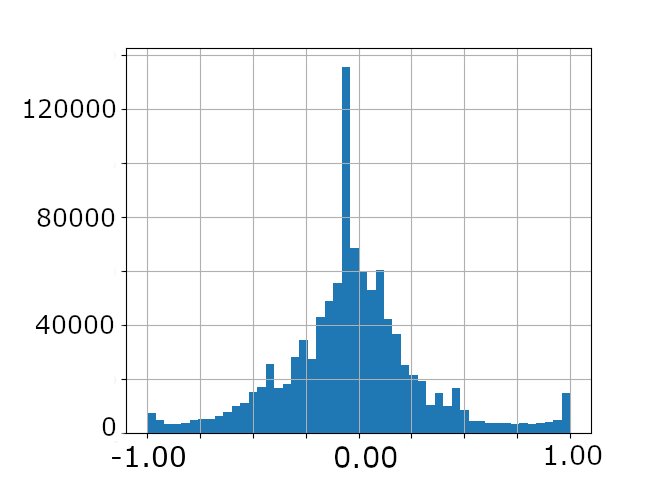}
    \caption{$\lambda=1$}
    \label{fig:lambda-1}
    \end{subfigure}
    ~
    \begin{subfigure}[t]{0.24\textwidth}
    \centering
    \includegraphics[width=1\linewidth, trim={0.5cm 0.5cm 0.5cm 0.5cm}, clip]{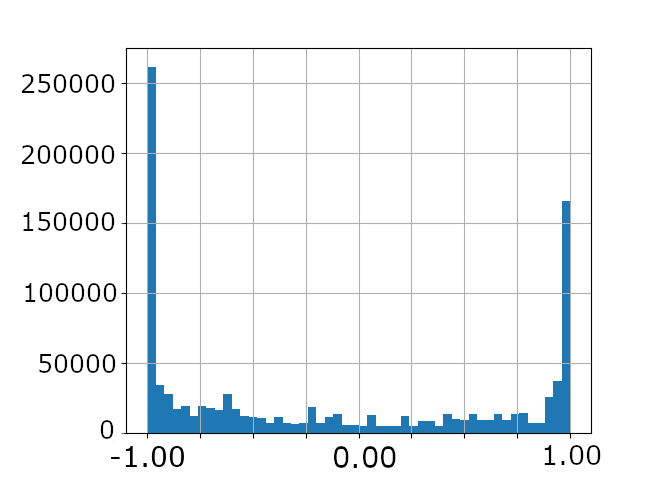}
    \caption{$\lambda=5$}
    \label{fig:lambda-5}
    \end{subfigure}
    ~
    \begin{subfigure}[t]{0.24\textwidth}
    \centering
    \includegraphics[width=1\linewidth, trim={0.5cm 0.5cm 0.5cm 0.5cm}, clip]{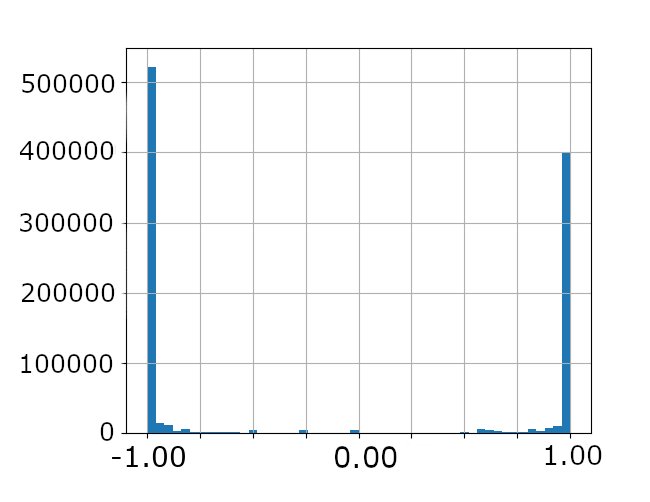}
    \caption{$\lambda=25$}
    \label{fig:lambda-25}
    \end{subfigure}
    ~
    \begin{subfigure}[t]{0.24\textwidth}
    \centering
    \includegraphics[width=1\linewidth, trim={0.5cm 0.5cm 0.5cm 0.5cm}, clip]{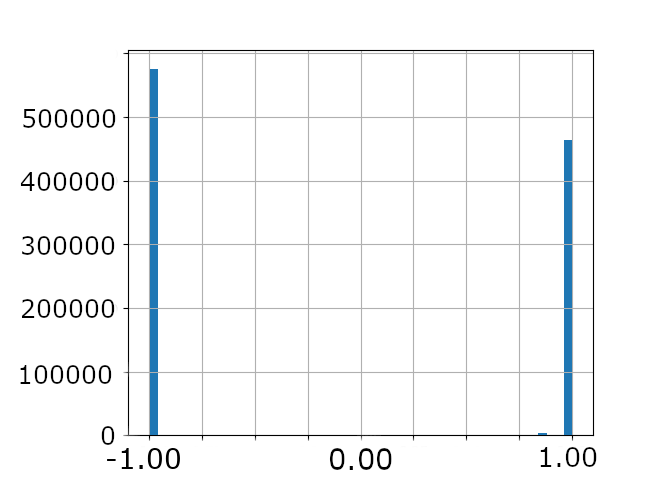}
    \caption{$\lambda=625$}
    \label{fig:lambda-625}
    \end{subfigure}

    \caption{Output distribution of $\tanh(\lambda x)$ for $\lambda=\{1,5,25,625\}$. Notice that starting with $\lambda=25$, in practice, the function behaves close to $\sgn(x)$.}
    \label{fig:lambda-quantization}
\end{figure*}

As $\lambda\to\infty$ the function converges to $\sgn(x)$. In a similar fashion, the derivative of the approximation function converges to the Dirac function $\delta$. In practice, as most of the features are outside of the region with high approximation error (see Fig.~\ref{fig:input-distribution}), we started observing close-to-binary results starting with $\lambda=25$. See Fig.~\ref{fig:lambda-quantization}.

In our tests we found that all the above approximation functions behaved similarly, however the best performance was obtained using the $\tanh$, while the softsign offered slightly lower performance. As such, the final reported results are obtained using the $\tanh$. During training we progressively increased the value of $\lambda$ starting from $2^0$ to $2^{16}$. 

\subsection{Stacked binary networks}\label{ssec:stacked-binary-nets}

As shown in~\cite{newell2016stacked}, using a stack of HG networks can be used to greatly improve human pose estimation accuracy, allowing the network to gradually refine its prediction at each stage. In a similar fashion, in this work we constructed a stack of binary HG networks also incorporating the improvements introduced in the previous subsections. We would like to verify to what extent stacking can further contribute on top of these improvements. In addition to these improvements, our method differs to~\cite{bulat2018hierarchical} in that all the intermediate layers used to join the stacks are also binarized. As the results from section~\ref{sec:results-human-pose} show, stacking further improves upon the improvements reported in the previous subsections.

\subsection{Combining binarization with distillation}\label{sec:distilled-binary-nets}

Recent work on knowledge distillation has focused on real-valued networks~\cite{hinton2015distilling}, largely ignoring the quantized, and especially, the binarized case. 

In this work, and in light of the methods proposed in the previous sub-sections, we also study the effect and effectiveness of knowledge distillation for the case of binary networks, evaluating in the process the following options: (a) using a real-valued teacher and a binary student and (b) using a binary teacher and a binary student with and without feature matching. During training, we used the output heatmaps of the teacher network as soft labels for the Binary Cross Entropy Loss. In addition, we found that the best results can be obtained by combining the ground truth and the soft labels with a weight equal to $0.25$.

\section{Human pose estimation experiments}\label{sec:results-human-pose}

In this section, we report our results on MPII, one of the most challenging datasets for single person human pose estimation~\cite{andriluka20142d}. MPII contains approximately 25,000 images and more than 40,000 persons annotated with up to 16 landmarks and visibility labels. We use the same split for validation and training as in~\cite{tompson2014joint} (3,000 for validation and 22,000 for training). We firstly report the performance improvements, using the PCKh metric~\cite{andriluka20142d}, obtained by applying \textbf{incrementally} the proposed methods in the same order as these methods appear in the paper. We then evaluate the proposed improvements in isolation.

\subsection{Results}
\paragraph{Baseline:} The performance of our strong baseline \cite{bulat2017binarized} using 1 HG with and without ReLU is shown in the first 2 rows of Table \ref{tab:results-binary-main}.\paragraph{Leaky non-linearities (Section~\ref{ssec:Symmetrical}):} The performance improvement obtained by replacing the ReLU with Leaky ReLU and then PReLU as proposed in our work is shown in the 3-rd and 4-th rows of Table~\ref{tab:results-binary-main}. We observe a large improvement of 2.5\% in terms of absolute error with the highest gains offered by the PReLU function. Note that we obtained similar accuracy between the variant that uses a single scale factor and the one that uses one per each channel for the negative slope. \paragraph{Reverse-order initialization (Section~\ref{ssec:progressive-initialisation}):} We observe an additional improvement of 0.8\% by firstly binarizing the features and then the weights, as proposed in our work and as shown in the 5-th row of Table \ref{tab:results-binary-main}. This, alongside the results from Fig.~\ref{fig:init-method} show that the proposed strategy is an efficient way of improving the performance of binary networks. \paragraph{Progressive binarization (Section~\ref{ssec:progressive-quantization}):} We observe an additional improvement of 0.4\% by the proposed progressive binarization as shown in the 6-th row of Table \ref{tab:results-binary-main}.\paragraph{Stacked binary networks (Section~\ref{ssec:stacked-binary-nets}):} We observe an additional improvement of 1.5\% and 1.9\% by using 2-stack and 3-stack HG networks, respectively, as shown in the 4-th column of Table~\ref{tab:results-binary-stacked}. While significant improvements can be observed when going from 1 HG to a stack of 2, the gain in performance diminishes when 1 more binary HG is added to the network. A similar phenomenon is also observed (but to less extent though) for the case of real-valued networks.

\paragraph{Binarization plus distillation}
As shown in the last row of Table \ref{tab:results-binary-main}, we obtained an improvement of 0.6\% via combining binarization and distillation for a binary network with a single HG distilled using a high performing real-valued teacher. Note that the binary network already incorporates the improvements proposed in section \ref{sec:binarization}. Also, the last column of Table~\ref{tab:results-binary-stacked} shows the improvements obtained by combining binarization with distillation for multi-stack binary HG networks. We observe an additional improvement of 1.5\% and 1.9\% by using 2-stack and 3-stack HG networks, respectively.

While we explored with using both a binary and a real-valued ``teacher'' given that finding a high performing binary teacher is challenging on its own, we obtained the best results using a real-valued one. However, in both cases the network converged to a satisfactory condition.

\begin{table}[!htbp]
\small
	\begin{center}
		\begin{tabular}{|c|c|c|c|c|c|}
			\hline
			Method  &  \begin{tabular}{@{}c@{}}Activ. \\ 
			Sec.~\ref{ssec:Symmetrical}\end{tabular} &  \begin{tabular}{@{}c@{}}Rev. init. \\ Sec.~\ref{ssec:progressive-initialisation}\end{tabular} & \begin{tabular}{@{}c@{}}Prog. bin. \\ Sec.~\ref{ssec:progressive-quantization}\end{tabular} & \begin{tabular}{@{}c@{}}Distill. \\ Sec.~\ref{sec:distilled-binary-nets}\end{tabular} & PCKh \\
			\hline\hline
			\cite{bulat2017binarized} & \xmark & \xmark & \xmark & \xmark & 76.6\% \\
			\cite{bulat2017binarized} & ReLU & \xmark & \xmark & \xmark & 76.3\% \\
			\hline 
			Ours & LReLU & \xmark & \xmark & \xmark & 78.1\% \\
			Ours & PReLU & \xmark & \xmark & \xmark & 79.1\% \\
			Ours & PReLU & \checkmark & \xmark & \xmark & 79.9\% \\
			Ours & PReLU & \checkmark & \checkmark & \xmark & 80.3\% \\
			Ours & \textbf{PReLU} & \textbf{\checkmark} & \textbf{\checkmark} & \textbf{\checkmark} & \textbf{80.9}\% \\
			\hline
            \cite{bulat2017binarized}-Real & - & - & - & - & 85.6\%   \\
            \hline
		\end{tabular}
	\end{center}
	\caption{PCK-h on the validation set of MPII for different configurations and methods. Each column points out to the section that presents or proposes the particular method.}
	\label{tab:results-binary-main}
\end{table}

\begin{table}[!htbp]
\small
	\begin{center}
		\begin{tabular}{|c|c|c|c|c|c|c|}
			\hline
			\#stacks & \#params & \cite{bulat2018hierarchical} & Ours w/o distil. & Ours w. distil \\
			\hline\hline
			  1 & 6.2M & 76.6\%  & 80.3\%&\textbf{80.9\%} \\
			  2 & 11.0M & 79.9\% & 81.8\% & \textbf{82.3\%} \\
			  3 & 17.8M & 81.3\%  & 82.2\% & \textbf{82.7\%} \\
			\hline
		\end{tabular}
	\end{center}
	\caption{PCK-h on the validation set of MPII for various number of binarized stacked HGs.}
	\label{tab:results-binary-stacked}
\end{table}

While the above results illustrate the accuracy gains by incrementally applying our improvements it is also important to evaluate the performance gains of each proposed improvement in isolation. As the results from Table~\ref{tab:results-binary-individual} show, the proposed techniques also yield high improvements when applied independently. At the same time, when evaluated in isolation the proposed modification offers a noticeable higher performance increase  compared with the case where they are gradually added together (i.e 0.8\% vs 1.9\% for reverse-order initialization).

\begin{table}[!htbp]
\small
	\begin{center}
		\begin{tabular}{|c|c|c|c|c|c|}
			\hline
			Method  &  \begin{tabular}{@{}c@{}}Activ. \\ 
			Sec.~\ref{ssec:Symmetrical}\end{tabular} &  \begin{tabular}{@{}c@{}}Rev. init. \\ Sec.~\ref{ssec:progressive-initialisation}\end{tabular} & \begin{tabular}{@{}c@{}}Prog. bin. \\ Sec.~\ref{ssec:progressive-quantization}\end{tabular} & \begin{tabular}{@{}c@{}}Distill. \\ Sec.~\ref{sec:distilled-binary-nets}\end{tabular} & PCKh \\
			\hline\hline
			\cite{bulat2017binarized} & \xmark & \xmark & \xmark & \xmark & 76.6\% \\
			\cite{bulat2017binarized} & ReLU & \xmark & \xmark & \xmark & 76.3\% \\
			\hline 
			Ours & LReLU & \xmark & \xmark & \xmark & 78.1\% \\
			Ours & PReLU & \xmark & \xmark & \xmark & 79.1\% \\
			Ours & \xmark & \checkmark & \xmark & \xmark & 78.5\% \\
			Ours & \xmark & \xmark & \checkmark & \xmark & 78.0\% \\
			Ours & \xmark & \xmark & \xmark & \checkmark & 77.6\% \\
			\hline
		\end{tabular}
	\end{center}
	\caption{PCK-h on the validation set of MPII that evaluates each proposed improvement in isolation. Each column points out to the section that presents the particular method.}
	\label{tab:results-binary-individual}
\end{table}

\subsection{Training}

We trained all models for human pose estimation (both real-valued and binary) following the same procedure: they were trained for 120 epochs, using a learning rate of $2.5e-4$ that was dropped every 40 epochs by a factor of 10. For data augmentation, we applied random flipping, scale ($0.75\times$ to $1.25\times$) jittering and rotation ($-30^{\circ}$ to  $30^{\circ}$). Instead of using the MSE loss, we followed the findings from~\cite{bulat2017binarized} and used the BCE loss defined as:
\begin{equation}
    \small
	l_{h} = \frac{1}{N}\sum_{n=1}^{N}\sum_{i=1}^{W}\sum_{j=1}^{H}p_{ij}^{n}\log{\widehat{p}_{ij}^{n}} + (1 - p_{ij}^{n})\log{(1 - \widehat{p}_{ij}^{n})},
\end{equation}
where $p_{ij}^{n}$ denotes the ground truth confidence map of the $n-$th part at pixel location $(i,j)$ and $\widehat{p}_{ij}^{n}$ is the corresponding predicted output at the same location. For distillation, we simply applied a BCE loss using as ground truth the predictions of the teacher network. 

\begin{figure}[!h]
    \centering
    \includegraphics[width=1.0\linewidth, trim={0.3cm 0.3cm 0.3cm 0.3cm}, clip]{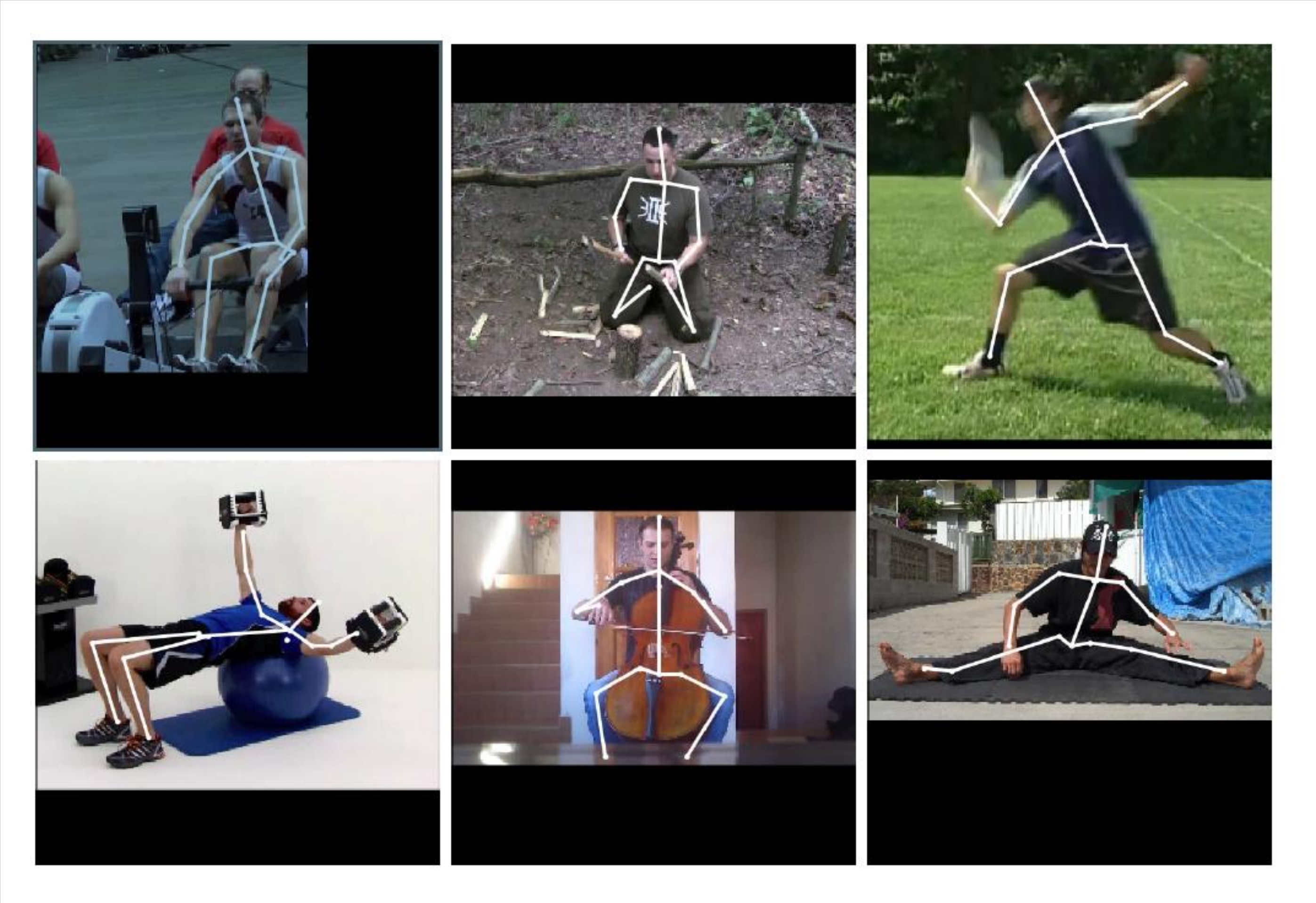}
    \caption{Qualitative results produced by our binary network on the validation set of MPII.}
    \label{fig:examples-human}
\end{figure}

The models were optimized using RMSProp~\cite{tieleman2012lecture}. We implemented our models using PyTorch~\cite{paszke2017automatic}. Qualitative results produced by our best-performing binary model can be seen in Fig.~\ref{fig:examples-human}.

\section{Imagenet classification experiments}\label{sec:results-imagenet}

To emphasize the generalization properties of the proposed improvements: (a)~leaky non-linearities (Section~\ref{ssec:Symmetrical}), (b)~reverse-order initialization (Section~\ref{ssec:progressive-initialisation}), (c)~smooth progressive quantization (Section~\ref{ssec:progressive-quantization}), and (d)~knowledge distillation (Section~\ref{sec:distilled-binary-nets}), in this section, we show that they are largely task-, architecture- and block-independent, by applying them on both of a more traditional architecture (i.e AlexNet~\cite{krizhevsky2012imagenet}) and a resnet-based one (ResNet-18) for the task of  Imagenet~\cite{deng2009imagenet} classification.

\paragraph{AlexNet:} Similarly to \cite{rastegari2016xnor, courbariaux2016binarized}, we removed the local normalization layer preserving the same structure, namely (from input to output): $C[3,96,(11\times11),4]$, $C[96,256,(5\times5),1]$, $C[256,384,(3\times3),1]$, $C[384,384,(3 \times 3),1]$, $C[384,256,(3\times3),1]$, $L(4096,4096)$, $L(4096,4096)$, $L(4096,1000)$, applying max-pooling after the $1-$st, $2-$nd and $5-$th layers. Similarly to~\cite{rastegari2016xnor}, the first and last layer were kept real.

\paragraph{ResNet:} We kept the original ResNet-18 macro-architecture unchanged, using, as proposed in~\cite{rastegari2016xnor}, the basic block from~\cite{he2016identity} with pre-activation. Similarly to the previous works~\cite{rastegari2016xnor}, the first and last layers were kept real.

\paragraph{Results:} As Table~\ref{tab:results-binary-imagenet} shows, when compared against the state-of-the-art method of~\cite{rastegari2016xnor} and~\cite{courbariaux2016binarized}, our approach offers a large improvement of up to ~4\% in terms of absolute error for both Top-1 and Top-5 error metrics using both AlexNet and ResNet-18 architectures. This further validate the generality of our method.

\begin{figure*}[!htbp]
    \centering
    \begin{subfigure}[t]{0.47\textwidth}
    \centering
    \includegraphics[width=0.73\linewidth]{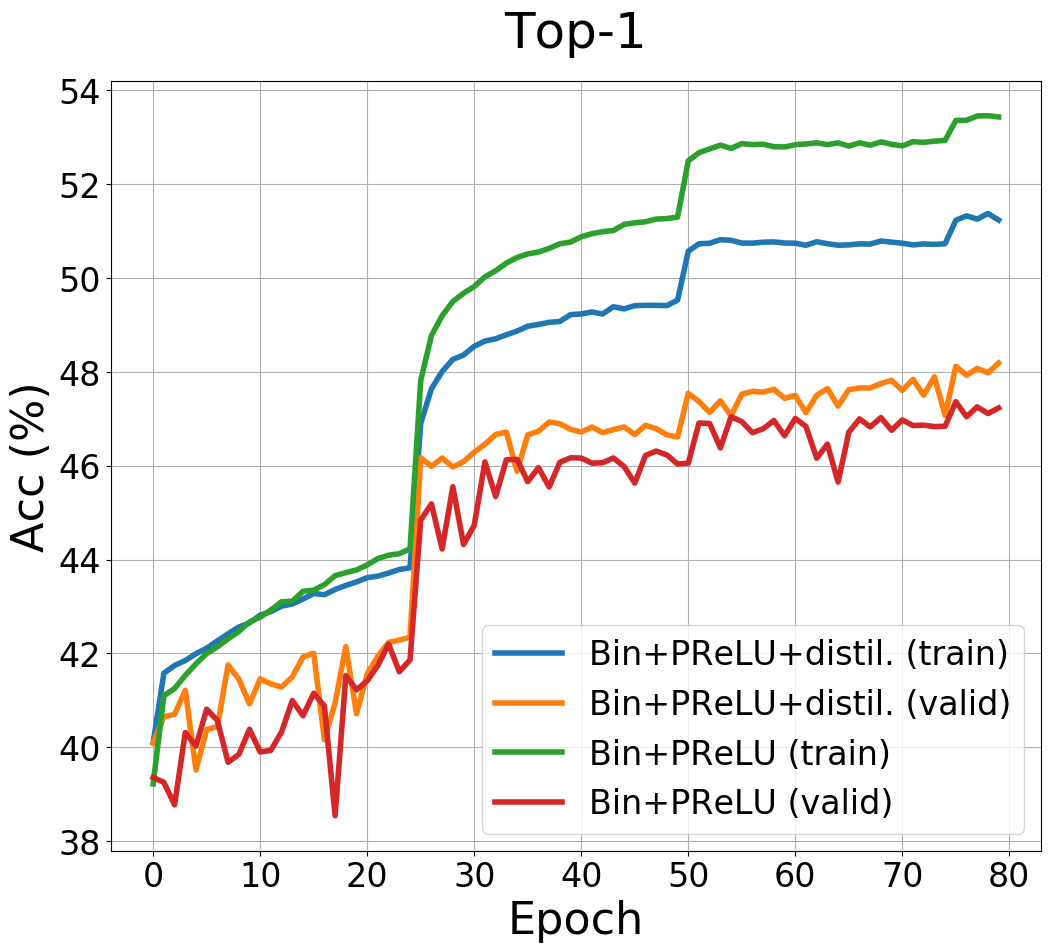}
    \caption{Top-1 accuracy on ImageNet.}
    \label{fig:imagenet-convergence-top1}
    \end{subfigure}
    ~
    \begin{subfigure}[t]{0.47\textwidth}
    \centering
    \includegraphics[width=0.73\linewidth]{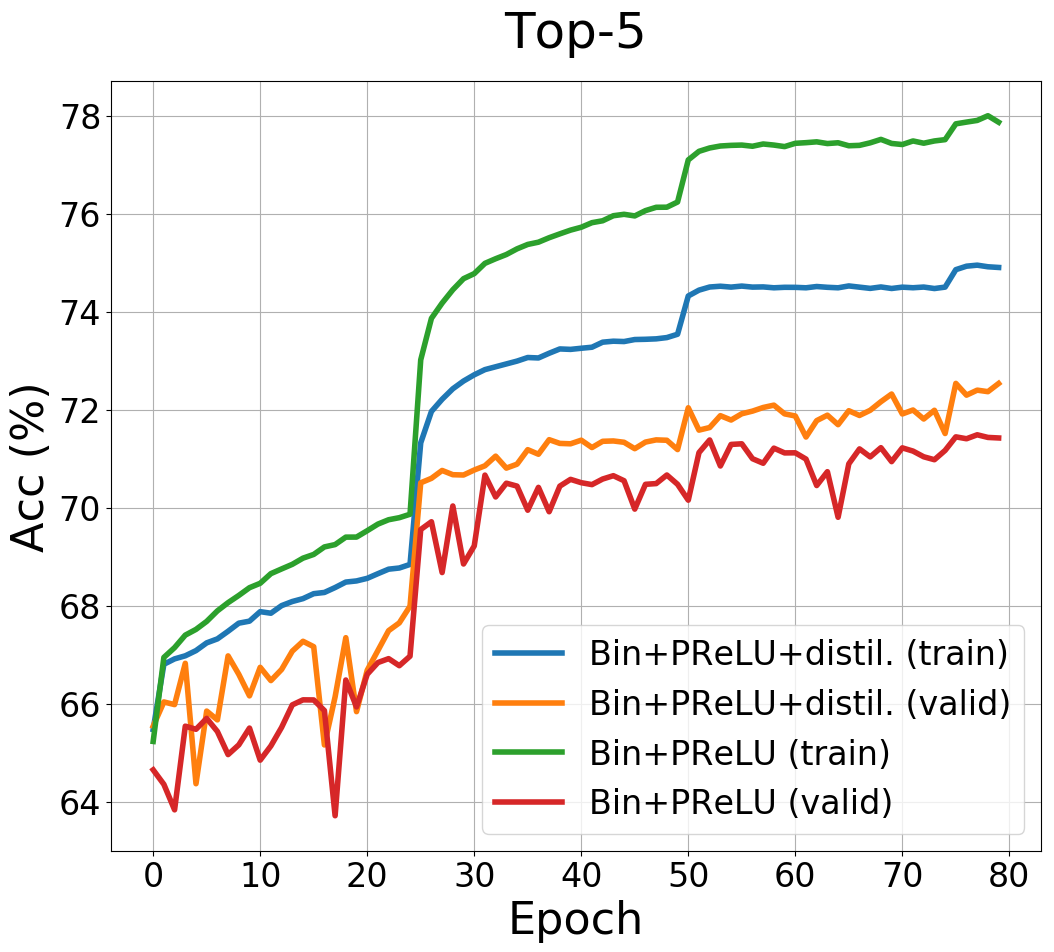}
    \caption{Top-5 accuracy on ImagenNet.}
    \label{fig:imagenet-convergence-top5}
    \end{subfigure}
    \caption{ImageNet training and validation accuracy vs epoch for different variants of our binary AlexNet.}
    \label{fig:imagenet-convergence}
\end{figure*}

\begin{table*}[!htbp]
	\begin{center}
		\begin{tabular}{|l|c|c|c|c|c|}
			\hline
			\multicolumn{5}{|c|}{Classification accuracy (\%)} \\
			\hline
			\multicolumn{1}{|c|}{Method} & \multicolumn{2}{c|}{AlexNet} & \multicolumn{2}{c|}{ResNet-18}\\
			\cline{2-5}
			 & Top-1 accuracy & Top-5 accuracy & Top-1 accuracy & Top-5 accuracy \\
			\hline\hline
			  BNN~\cite{courbariaux2016binarized} & 41.8\% & 67.1\% & 42.2\% & 69.2\%  \\
			  XNOR-Net~\cite{rastegari2016xnor} & 44.2\% & 69.2\% & 51.2\%& 73.2\% \\
			  \textbf{Ours} & \textbf{48.6\%} & \textbf{72.8\%} & \textbf{53.7}\% & \textbf{76.8}\%\\
			  \hline--expand-bbl
			  Real valued~\cite{krizhevsky2012imagenet} & 56.6\% & 80.2\% & 69.3\%& 89.2\%\\
			\hline
		\end{tabular}
	\end{center}
	\caption{Top-1 and Top-5 classification accuracy using binary AlexNet and ResNet-18 on Imagenet. Notice that our method offers consistent improvements across multiple architectures: both traditional ones(AlexNet) and residual ones (ResNet-18). }
	\label{tab:results-binary-imagenet} \vspace{10pt}
\end{table*}

\paragraph{Training:} We trained the binarized version of AlexNet~\cite{krizhevsky2012imagenet} and ResNet-18~\cite{he2016deep} using Adam~\cite{kingma2014adam} starting with a learning rate of $1e-3$ that is gradually decreased every 25 epochs by 0.1. We simply augment the data by firstly resizing the images to have 256px over the smallest dimension and then randomly cropping them to $227\times227$ for AlexNet and $224\times224$px for ResNet. We believe that further performance gains can be achieved with more aggressive augmentation. At test time, instead of random-crop we center-crop the images. To alleviate problems introduced by the binarization process, and similarly to~\cite{rastegari2016xnor}, we trained the network using a large batch size, specifically 400 for AlexNet and 256 for ResNet-18. All models were trained for 80 epochs. Fig.~\ref{fig:imagenet-convergence} shows the top-1 and top-5 accuracy across training epochs for AlexNet (the network was initialized using the procedure proposed in Section~\ref{ssec:progressive-initialisation}).

\section{Conclusions}

In this work, we proposed a series of novel techniques for highly efficient binarized convolutional neural network. We experimentally validated our results on the challenging problems of human pose estimation and large scale image classification. Mainly, we propose (a) more appropriate non-linear activation functions, (b) reverse-order initialization, (c) progressive features quantization, and (d) network stacking that improve existing state-of-the-art network binarization techniques. Furthermore, we explore the effect and efficiency of knowledge distillation procedures in the context of binary networks using a real-valued ``teacher'' and binary ``student''.

Overall, our results show that a performance improvement of up to 5\% in absolute terms is obtained on the challenging human pose estimation dataset of MPII. Finally, we show that our approach is architecture and task-agnostic and can increase the performance of arbitrary networks. In particular, by applying the proposed techniques to Imagenet classification, we report an absolute performance improvement of 4\% over the current state-of-the-art using both AlexNet and ResNet architectures.

{\small
\bibliographystyle{ieee}

}

\end{document}